%% file: paper.tex
\documentclass[conference]{IEEEtran}
\IEEEoverridecommandlockouts
\usepackage{cite}
\usepackage{amsmath,amssymb,amsfonts}
\usepackage{algorithmic}
\usepackage{graphicx}
\usepackage{textcomp}
\usepackage{xcolor}
\def\BibTeX{{\rm B\kern-.05em{\sc i\kern-.025em b}\kern-.08em
    T\kern-.1667em\lower.7ex\hbox{E}\kern-.125emX}}
\newcommand{\pomdp}{\textsc{pomdp}}

\newcommand{\St}{\mathcal{S}}
\newcommand{\Ac}{\mathcal{A}}
\newcommand{\Pd}{\mathcal{P}}
\newcommand{\Ob}{\mathcal{O}}

\usepackage{mathtools}
\DeclareMathOperator*{\E}{\mathbb{E}}
\newcommand{\R}{\mathbb{R}}

\DeclareMathOperator*{\argmax}{arg\,max}

\DeclarePairedDelimiterX{\expectarg}[2]{\{}{\}}{%
  #1\,\delimsize\vert\,\mathopen{}#2}
\newcommand\MTkillspecial[1]{
  \bgroup
  \catcode`\&=9
  \let\\\relax%
  \scantokens{#1}%
  \egroup
}
\DeclarePairedDelimiter\myabs\lvert\rvert
\reDeclarePairedDelimiterInnerWrapper\myabs{star}{
  \mathopen{#1\vphantom{\MTkillspecial{#2}}\kern-\nulldelimiterspace\right.}
  #2
  \mathclose{\left.\kern-\nulldelimiterspace\vphantom{\MTkillspecial{#2}}#3}}
\DeclarePairedDelimiter\mybracket{\{}{\}}
\reDeclarePairedDelimiterInnerWrapper\mybracket{star}{
  \mathopen{#1\vphantom{\MTkillspecial{#2}}\kern-\nulldelimiterspace\right.}
  #2
  \mathclose{\left.\kern-\nulldelimiterspace\vphantom{\MTkillspecial{#2}}#3}}
\usepackage{hyperref}
\usepackage{booktabs}
\usepackage{tikz}
\usetikzlibrary{shapes.geometric,shapes.arrows,decorations.pathmorphing}
\usetikzlibrary{matrix,chains,scopes,positioning,arrows,fit,arrows.meta}
\usetikzlibrary{calc}
\newcounter{num}
\newcommand{\tictactoe}[1] {
  \begin{tikzpicture}[line width=0.6pt, scale=0.4, remember picture]
    \def\r{0.8mm}
    \tikzset{
      circ/.pic={\draw circle (\r);},
      cross/.pic={\draw (-\r,-\r) -- (\r,\r) (-\r,\r) -- (\r,-\r);},
      opt/.pic={\draw[opacity=0.2] (-\r,-\r) -- (\r,\r) (-\r,\r) -- (\r,-\r);}
    }

    \foreach \i in {1,2} \draw (\i,0) -- (\i,3) (0,\i) -- (3,\i);

    \setcounter{num}{0}
    \foreach \y in {0,...,2}
    \foreach \x in {0,...,2}
      {
        \coordinate (\thenum) at (\x+0.5,2-\y+0.5);
        \addtocounter{num}{1}
      }

    \coordinate (tttbl) at (0, 0);
    \coordinate (tttbr) at (3, 0);
    \coordinate (ttttl) at (0, 3);
    \coordinate (ttttr) at (3, 3);

    \def\X{X} \def\x{x} \def\O{O} \def\n{n}

    \foreach \l [count = \i from 0] in {#1}
      {
        \if\l\X \path (\i) pic{cross};
        \else
          \if\l\O \path (\i) pic{circ};
          \else
            \if\l\x \path (\i) pic{opt};
            \else
              \if\l\n \node[opacity=0.5] at (\i) {\sffamily\i};
              \fi
            \fi
          \fi
        \fi
      }
  \end{tikzpicture}
}
\begin{document}

\title{Towards Using Fully Observable Policies for \textsc{pomdp}s
  \thanks{Supported by the ÚNKP-21-3 New National Excellence Program of the
    Ministry for Innovation and Technology from the source of the National
    Research, Development and Innovation Fund.}
}

\author{\IEEEauthorblockN{András Attila Sulyok}
\IEEEauthorblockA{\textit{Faculty of Information Technology and Bionics} \\
\textit{Pázmány Péter Catholic University}\\
Budapest, Hungary \\
{sulyok.andras.attila@itk.ppke.hu}}
\and
\IEEEauthorblockN{Kristóf Karacs}
\IEEEauthorblockA{\textit{Faculty of Information Technology and Bionics} \\
\textit{Pázmány Péter Catholic University}\\
Budapest, Hungary \\
{karacs@itk.ppke.hu}}
}

\maketitle

\begin{abstract}
  Partially Observable Markov Decision Process (\textsc{pomdp}) is a framework
  applicable to many real world problems. In this work, we propose an approach
  to solve \textsc{pomdp}s with multimodal belief by relying on a policy that
  solves the fully observable version. By defininig a new, mixture value
  function based on the value function from the fully observable variant, we
  can use the corresponding greedy policy to solve the \textsc{pomdp} itself.
  We develop the mathematical framework necessary for discussion, and introduce
  a benchmark built on the task of Reconnaissance Blind TicTacToe. On this
  benchmark, we show that our policy outperforms policies ignoring the
  existence of multiple modes.
\end{abstract}


\input{introduction.tex}

\section{Methods}

\input{maths.tex}

\input{methods.tex}

\input{experiments.tex}

\input{conclusion.tex}


\bibliographystyle{IEEEtran}
\IEEEtriggeratref{9}


\end{document}

%% file: introduction.tex
\section{Introduction}
\label{sec:introduction}

Markov Decision Process (\textsc{mdp}) is a general framework that can be used
in many practical applications where there is an agent that is in interaction
with its environment, and there is a reward function that the agent needs to
maximize \cite{Sutton2018}. Numerous algorithms have been proposed and are in
use to effectively solve \textsc{mdp}s (for example, \cite{mnih2015, ppo, mpo}).

On the other hand, \textsc{mdp}s assume that the agent knows the state the
environment is in at any time, i.e., all information to plan ahead is available.
Partially Observable Markov Decision Process (\pomdp{}, \cite{pomdp_control}) is
a modification to the \textsc{mdp} theory that includes partial visibility:
instead of observing the state directly, it is observed indirectly via an
observation function. This framework is applicable in many areas from autonomous
driving to healthcare \cite{Komorowski2018} or education \cite{Mandel2014}.

One way to deal with the uncertainty introduced by partial observability is to
keep track of the posterior probability of the current state given the
observation and action history, i.e., the \emph{belief} \cite{Hauskrecht2000}.
In this work, we focus on a specific class of \pomdp{}s, inspired by the recent
challenge of Reconnaissance Blind Chess (\textsc{rbc}, \cite{Markowitz2018,
  newman2016reconnaissance, rbc_conclusion}), in which the belief is a
multimodal distribution, that is, it is not enough to keep track of only one
state, which is presumed to be a corrupted version of the true underlying state
of the environment, but instead there are multiple distinct states with too high
probability to be simply ignored. We assume that the belief can be calculated.

We also assume that we have access to a (reasonably good) policy that solves the
fully observable case, that is, the same environment but with the observation
function being the identity function. For many applications, this is not an
unreasonable assumption, for example, one can build a simulator for the agent
and do training there (see, for example \cite{openai_rubik}).

Based on these, we propose a mixture policy to solve \pomdp{}s that uses the
policy for the fully observable case weighted by the belief. We introduce the
necessary mathematical framework for this, and show its performance on a
benchmark that is a variation of \textsc{rbc} called Reconnaissance Blind
TicTacToe.

\subsection{Related work}

The traditional approach to solving \pomdp{}s is to transform the problem into
belief-\textsc{mdp}s \cite{Hauskrecht2000, Kurniawati2008, Jaakkola1995}, which
are \textsc{mdp}s that use the beliefs of the original problem as their state
space. Since this means an exponentially large state space, various estimation
techniques were developed \cite{Kurniawati2008}; however, these are hard to
generalize for continuous state spaces.

There are several attempts to solve \pomdp{}s using deep reinforcement learning
techniques, by observation aggregation \cite{mnih2015} or recurrent models
\cite{wierstra2007, Hausknecht2015}, however, these were not specifically
designed for tasks involving multimodal belief. This can result in undesired
linear combinations of modes, i.e. instead of keeping track of probable states,
the model might keep track of states of lower probability instead, corresponding
to combinations of two or more modes.

There are also works proposing to solve a \pomdp{} using a policy for the fully
observable case, especially for the \textsc{rbc} challenge \cite{Clark2021,
  Blowitski2021, Highley2020, Guo2020}, however, these do not give a clear
mathematical description and do not discuss their solutions in terms of general
\pomdp{}s. They also focus more on the speciality of \textsc{rbc} that the
location of the observation is directly controlled by the agent and on how to
limit the number of probable states.


%% file: maths.tex
\subsection{\pomdp{} background}

In general, \pomdp{}s \cite{pomdp_control} are defined by the tuple $\langle
  \St, \Ac, \Ob, \Pd, O, R, \gamma \rangle$, where $\St$ denotes the (unobservable)
states, $\Ac$ the set of actions, $\Ob$ the set of observations and $R$ the
reward function. We denote the
state, action, reward and observation at timestep $t$ as the random variables
$S_t \in \St$, $A_t \in \Ac$, $R_t \in \R$ and $O_t \in \Ob$, respectively. The
dynamics is governed by the transition distribution $\Pd(s', r \mid s, a) =
  \Pr(S_{t+1} = s', R_{t+1} = r \mid S_t = s, A_t = a)\; (\forall t)$ and the
observation function $O(o \mid s) = \Pr(O_t = o \mid S_t = s)$.

To simplify notation, we denote observations up to timestep $t$ as $o_{\le t} =
  (o_0, \dots, o_t)$, and actions before timestep $t$ as $a_{< t} = (a_0, \dots,
  a_{t-1})$.

A policy is a distribution over actions: $\pi(a \mid o_{\le t}, a_{< t}) =
  \Pr(A_t = a \mid o_{\le t}, a_{< t})$ given the history of past observations
and actions $o_0, a_0, \dots, o_t$.  The goal of a \pomdp{} is to find a
policy so that the expected cumulative reward is maximal:
$\argmax_\pi \E_{(S_t, A_t, R_{t+1}) \sim \Pd, \pi} \sum_t \gamma^t
R_{t+1}$.\footnote{To simplify discussion, we will assume that the discount
  factor $\gamma = 1$. The generalization to the discounted case is
  straightforward.}

We denote the belief over possible states as $\beta_t = \Pr(S_t \mid o_{\le t},
  a_{<t})$.

  Following the \pomdp{} literature \cite{Hauskrecht2000}, we define the
\emph{belief-\textsc{mdp}} of a \pomdp{} as the tuple $\langle B, \Ac,
  \widetilde{\Pd}, \widetilde{R}, \gamma \rangle$, where $B = \Delta^\St$ is
the state space of beliefs, $\widetilde{R}$ is the reward function:
\begin{equation}
  \widetilde{R}(\beta) = \E_{s \sim \beta} R(s).
\end{equation}

The dynamics of the belief-\textsc{mdp} is governed by a transition function of
the beliefs: after issuing an action, the agent observes a new observation, and
transitions into a new belief-state. The distribution of the next observation
$o_{t+1}$, given current belief $\beta_t$ and action $a_t$ is:
\begin{IEEEeqnarray}{rCl}
  \IEEEeqnarraymulticol{3}{l}{p(o_{t+1} \mid \beta_t, a_t)} \notag \\*
  \IEEEyesnumber\IEEEyessubnumber*
  \quad
  & =& \sum_{s_t} \beta_t(s_t) \sum_{s_{t+1}}
  \Pd(s_{t+1} \mid s_t, a_t) O(o_{t+1} \mid s_{t+1})                                                          \\
  & =& \E_{s_t \sim \beta_t} \E_{s_{t+1} \mid \Pd(\cdot \mid s_t, a_t)} O(o_{t+1}
  \mid s_{t+1}).
\end{IEEEeqnarray}

After the observation, the belief-transition is deterministic:
\begin{IEEEeqnarray}{rCl}
  \IEEEeqnarraymulticol{3}{l}{
    \beta_{t+1}(\beta_t, a_t, o_{t+1})(s_{t+1})} \notag\\*
  \IEEEyesnumber\IEEEyessubnumber*
  \quad
  & =& p(s_{t+1} \mid \beta_t, a_t, o_{t+1})                   \\
  & =& \E_{s_t \sim \beta_t} p(s_{t+1} \mid s_t, a_t, o_{t+1}) \\
  & =& \E_{s_t \sim \beta_t} \left\{
  \frac{p(o_{t+1} \mid s_{t+1}, s_t, a_t)p(s_{t+1} \mid s_t, a_t)}{p(o_{t+1}
    \mid s_t, a_t)}
  \right\}                                                     \\
  & =& \E_{s_t \sim \beta_t} \left\{
  \frac{p(o_{t+1} \mid s_{t+1})p(s_{t+1} \mid s_t, a_t)}
  {\sum_{s_{t+1}'} p(o_{t+1} \mid s_{t+1}') p(s_{t+1}' \mid s_t, a_t)}
  \right\}                                                     \\
  & =& \E_{s_t \sim \beta_t} \left\{
  \frac{O(o_{t+1} \mid s_{t+1})\Pd(s_{t+1} \mid s_t, a_t)}
  {\sum_{s_{t+1}'}O(o_{t+1} \mid s_{t+1}')\Pd(s_{t+1}' \mid s_t, a_t)}
  \right\}. \label{eq:belief_update}
\end{IEEEeqnarray}



%% file: methods.tex
\subsection{One-step estimation of the Q function}

Similarly to the value function defined for an \textsc{mdp}:
\begin{equation}
  Q^\pi(s, a) = \E_{s_t \sim \Pd, a_t \sim \pi}
  \expectarg*{\sum_{t=0}^T R(s_t)}{s_0 = s, a_0 = a},
\end{equation}
we can define the value function for \pomdp{}s:
\begin{equation}
  Q^\pi(\beta, a) = \E_{\beta_t \sim \widetilde{\Pd}, a_t \sim \pi}
  \expectarg*{\sum_{t=0}^T \widetilde{R}(\beta_t)}{\beta_0 = \beta, a_0 = a}.
\end{equation}

Since calculating $Q(\beta, a)$ is NP-hard in general, there are different
methods to estimate it \cite{Hauskrecht2000,Kurniawati2008}. We propose a
different approach.

As we described in Section~\ref{sec:introduction}, we assume we have a (reasonably
good) policy $\pi_o$ for the fully observable case and a corresponding Q
function $Q^{\pi_o}$. (Note that this is not the belief-\textsc{mdp} of the
\pomdp{}, but the \textsc{mdp} with the same dynamics and reward, just without
the observation.)

This assumption is reasonable in a lot of applications, because solving the
fully observable case is significantly easier. In a lot of cases, there are even
procedural or analytic solutions. In absence of this, if the dynamics are known
in a form that is easy to sample from, one can build a simulator and train a
policy there \cite{Sutton2018}.

Our other assumption is that the belief $\beta_t$ is multi-modal: it cannot be
modelled as a mean value and some additive noise. This means simply guessing
what the most probable state is at each timestep, and making a decision based on
that is not enough.

Instead, we propose to define a \pomdp{} policy $\pi_{po}$ based on the one-step
estimation of the Q function:
\begin{equation}\label{eq:mixture_policy}
  \widetilde{Q}^{\pi_{o}}(\beta, a) := \E_{s
    \sim \beta} Q^{\pi_o}(s, a).
\end{equation}

Based on this definition, the \pomdp{} policy can be defined as the
greedy policy with respect to $\widetilde{Q}^{\pi_{o}}$:
\begin{equation}
  \pi_{po}(a \mid \beta) := \begin{cases}
    1 & \text{if } a = \argmax_{a' \in a} \widetilde{Q}^{\pi_o}(\beta, a')\\
    0 & \text{otherwise}
  \end{cases}
\end{equation}
(if we assume the maximum occurs for a single action or we choose one of the
maximizing actions).

We call this estimation one-step, because, since $\pi_{po} \neq \pi_{o}$,
$Q^{\pi_{o}}$ does not accurately represent the expected reward in the future,
and we deal with the uncertainty coming from partial observability only in
timestep $t$ when we take the expectation over $\beta_t$. In other words,
\(Q^{\pi_{po}}(\beta, a) \neq \widetilde{Q}^{\pi_o}(\beta, a)\), even if they
might be close for a certain problem class.

However, there is intuitive motivation behind this scheme, as it is the
mathematical formulation of the question ``What would a policy do that can see
the state?'', weighted by our belief (probability) of each state.

Care must be taken when using a learned $Q^{\pi_{o}}$: usually, reinforcement
learning algorithms learn from interactions between the agent and the
environment, and the accuracy of information depends on which states (and
actions) the agent visited during training. In particular, there might be states
that the agent has never seen, and for which the Q function is uninitialized.

This is normally not a problem if the same policy is used for training and
inference, or when the Q function is represented by powerful function
approximators capable of generalization, but in \eqref{eq:mixture_policy},
$Q^{\pi_o}$ must have valid values for every state with $\beta_t(s) > 0$.

Another potential way to deal with this is to start the simulator from different
states the agent can potentially reach.


%% file: experiments.tex
\section{Experiments}

\subsection{Metrics used}
\label{sec:metrics}

To be able to analyze the fully-observable policy-based method, we used several
metrics, the most straightforward of which is the average cumulated reward for
each episode (the \pomdp{} objective).

We used a ``proxy policy'' for every alternative policy that does not use the
complete belief information: at each step, we chose the states with maximal
belief:
\begin{equation}
  S^{(t)}_{\text{max}} = \left\{ s \in \St : \beta_t(s) = \max_{s'} \beta_t(s')
  \right\},
\end{equation}
taking into account that multiple states might be maximizing.

In this section, for notational convenience, we will implicitly assume that
there may be multiple elements that maximize an $\argmax$ operator, so we define
the $\argmax$ operator to be the set of maximizing elements:
\begin{equation}
  \argmax_{x \in S} f(x) := \left\{ x \in S : f(x) = \max_{x'} f(x') \right\}.
\end{equation}

Note that the assumption for this proxy policy is that it calculates the belief
$\beta_t$ for each timestep, but then \emph{discards} information, retaining
only the states with maximal probability. The intuition behind this is that the
efficiency of this functions as an upper bound on any policy taking action based
on any sort of state-identification method.

We compared how the actions taken by the alternative policy differ from our
proposed policy: we computed the intersection-over-union (IoU) of the maximizing
actions for each timestep. Formally, for each timestep, we define the action set
from which the agent chooses one:
\begin{equation}
  A^{(t)}_{\text{mix}} = \argmax_{a \in \Ac} \widetilde{Q}(\beta_t, a)
  = \argmax_{a \in \Ac} \E_{s \sim \beta_t} Q(s, a)
\end{equation}
(``mix'' denoting that this is defined using a mixture of Q values)
and the probability-maximizing policy:
\begin{equation}
  A^{(t)}_{\text{max}} = \argmax_{a \in \Ac} \E_{s \sim \mathcal{U}\left(S^{(t)}_{\text{max}
      }\right)} Q(s, a),
\end{equation}
where $\mathcal{U}$ denotes the uniform distribution.
The IoU, or Jaccard index of these two action sets is:
\begin{equation}\label{eq:iou}
  \text{IoU} = \frac{|A_{\text{max}} \cap A_{\text{mix}}|}{|A_{\text{max}} \cup
  A_{\text{mix}}|}.
\end{equation}

Another metric we are interested in is the difference in value between the
actions chosen by the two policies. Ideally, we would compare using the true
value function $Q(\beta, a)$, but since that is hard to compute, we again
estimate it with the one step value function $\widetilde{Q}(\beta, a)$. That is,
we measure the value margin against the alternative (benchmark) policy:
\begin{equation}\label{eq:value_margin}
  M^{(t)}_{\text{alt}} = \widetilde{Q}\left(\beta_t, A^{(t)}_{\text{mix}}\right) -
  \E_{a \in \mathcal{U}\left(A^{(t)}_{\text{max}}\right)}
  \widetilde{Q}\left(\beta_t, a\right),
\end{equation}
where we can write $\widetilde{Q}(\beta_t, A^{(t)}_{\text{mix}})$ since the
value of $\widetilde{Q}$ is the same for all $a \in A^{(t)}_{\text{mix}}$ by
definition. Since this metric is based on the estimation $\widetilde{Q}(\beta,
  a)$ and not on the true $Q(\beta, a)$, comparing it to the difference in episode
returns can be indicative of the estimation error.


\subsection{Benchmark}
\label{sec:benchmark}

For benchmark, we used a Reconnaissance Blind TicTacToe (\textsc{rbt}, inspired
by Reconnaissance Blind Chess \cite{Markowitz2018, newman2016reconnaissance,
  rbc_conclusion}).

In this task, the agent plays a game of TicTacToe, except it cannot see the
moves of the opponent directly. Instead, before each move, a random subset of
the cells is revealed. In this document, we will refer to this as
\emph{sensing}; we used rectangle-shaped sensing windows of different sizes.

Reward is $1$ for winning, $-1$ for losing the game, $-1$ also
for an invalid move, and $0$ otherwise. In our implementation, invalid moves
also terminate the episode.

On the one hand, this task has the defining features of a ``hard'' \pomdp{}
problem: the observation is not simply a noisy estimation of the state the
agent: many possible (probable) but distinct states can be consistent with a
given observation history, and to retain all information, has to remember
observation potentially from the beginning of the episode.

\begin{figure*}[t]
  \centering
  \begin{tabular}{cc||ccccc}
    \tictactoe{
      , ,,
      ,X,,
      , ,,
    } \begin{tikzpicture}[remember picture, overlay]
        \draw[rounded corners, gray!70, thick] ($(tttbr)!1/3!(ttttl)$) rectangle
        (ttttl);
      \end{tikzpicture}          &
    \tictactoe{
      ,X,,
      ,X,O,
      , ,,
    }  \begin{tikzpicture}[remember picture, overlay]
         \draw[rounded corners, gray!70, thick] ($(tttbl)!1/3!(ttttr)$) rectangle (ttttr);
       \end{tikzpicture} &
    \tictactoe{
      O, X, ,
      , X, O,
      , , ,
    }                                                                                 &
    \tictactoe{
      ,X, ,
      O,X,O,
      , , ,
    }                                                                                 &
    \tictactoe{
      ,X, ,
      ,X,O,
      O, , ,
    }                                                                                 &
    \tictactoe{
      ,X, ,
      ,X,O,
      ,O, ,
    }                                                                                 &
    \tictactoe{
      ,X,  ,
      ,X, O,
      , , O,
    }                                                                                                                                              \\
    $o_1$                                                                             & $o_2$ & $0.125$ & $0.125$ & $0.25$ & $0.25$ & $0.25$       \\
                                                                                      &       &         &         &        &        & (true state)
  \end{tabular}
  \caption{Belief after $4$ plies ($\beta_2$) in a randomly chosen episode with
    sensing window size $2\times2$. On the left are the observations with
    sensing windows marked with gray rectangles, on the right are the states
    with positive probability. First row contains the state visualizations, with
    their respective probabilities below. The true (hidden) state is the
    rightmost one.}
  \label{fig:belief_visualisation}
\end{figure*}
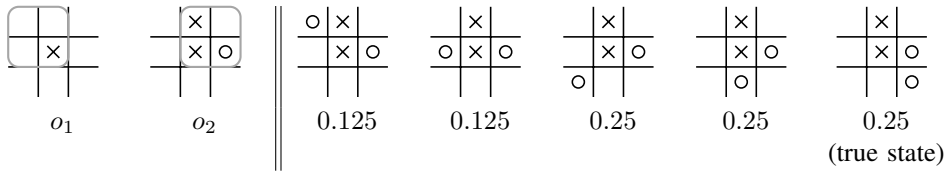

On the other hand, the task is small enough that the belief can be calculated
exactly using \eqref{eq:belief_update}; an example belief for $t=3$ is shown in
Fig.~\ref{fig:belief_visualisation}. This makes it possible to analyze the Q
mixture-based policy without any corruption from an imperfect belief-estimator.
A policy that solves the fully observable case is also easy to compute.


Note that the dynamics $\Pd(s_{t+1} \mid s_t, a_t)$ has two components: the
agent action, which is deterministic (given $a_t$), and the opponent action,
which is stochastic if the opponent policy is stochastic. Therefore, the
uncertainty (in the belief) comes from both the incomplete observation and the
stochastic dynamics, and by using different sense window sizes (controlling the
amount of information from observation) and different opponent policies, we can
control the amount and quality of the uncertainty in the environment.

There are two main differences from \textsc{rbc}: the opponent does not play
blind, i.e., it observes the board state fully. This removes the belief of the
opponent from the state \cite{Markowitz2018}, and frees the agent of any need to
bluff. The second change is that in \textsc{rbc}, invalid actions (actions
against the rules) resulted in no-operation actions, and since both players are
playing blind, neither would have any unfair advantage because of this. In
\textsc{rbt}, however, to simplify matters, we terminate the episode after an
invalid action. This means that the belief-update can implicitly assume the
action was valid, since otherwise the episode is terminated and the calculated
belief is not used anyway.

\subsection{Results}

\begin{figure}[tpb]
  \centering
  \includegraphics{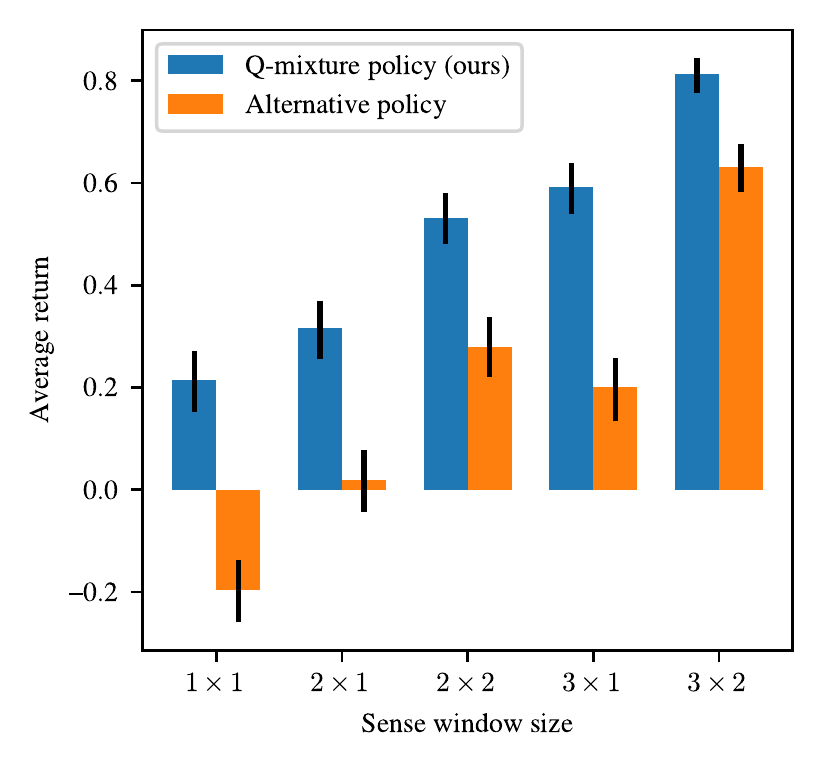}
  \caption{Average returns for the two policies across different sense window
    sizes. In this task, the reward is only given at the end of the episode, $1$
    for winning and $-1$ for losing, hence it is proportional to the win ratio.
    The measurements are based on $1000$ episodes. Error bars indicate the
    $95\%$ confidence estimates. Numerical values are shown in
    Table~\ref{tab:average_returns}.}
  \label{fig:average_returns}
\end{figure}

Fig.~\ref{fig:average_returns} and Table~\ref{fig:average_returns} show the
average returns of our proposed policy, with the alternative policy that only
takes the state with maximum probability into account (as described in
Section~\ref{sec:metrics}). As shown by the figure, our proposed policy
consistently outperforms the alternative.

To model the varying amount of information available in the observation, and
with that, the number of probable states, we ran the experiment with different
sensing window sizes. As expected, the larger the sensing window, the more
information the model has, and the better the performance of both policies
overall.

\begin{table}
  \centering
  \caption{Average returns for the two policies across different sense window
    sizes corresponding to Fig.~\ref{fig:average_returns}.}
  \label{tab:average_returns}
  \begin{tabular}{ccr@{${}\pm{}$}l}
    \toprule
    Sense window & Q-mixture policy (ours) & \multicolumn{2}{c}{Alternative
    policy}                                                                          \\
    \midrule
    $1\times1$   & $0.215\pm0.06$          & $-0.197$   & $0.06$ \\
    $2\times1$   & $0.316\pm0.06$          & $0.019$    & $0.06$ \\
    $2\times2$   & $0.532\pm0.06$          & $0.279$    & $0.06$ \\
    $3\times1$   & $0.592\pm0.06$          & $0.2$      & $0.06$ \\
    $3\times2$   & $0.813\pm0.05$          & $0.632$    & $0.05$ \\
    \bottomrule
  \end{tabular}
\end{table}

\begin{figure}[tpb]
  \centering
  \includegraphics{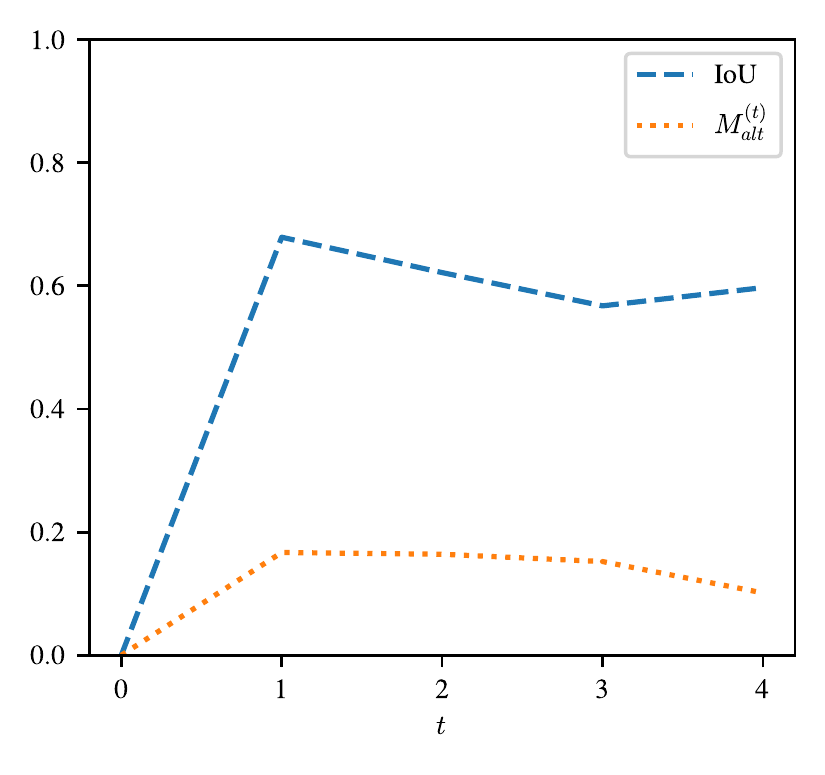}
  \caption{Metrics described in Section~\ref{sec:metrics} across time in an
    episode with $2\times2$ sense window. ``IoU'' is defined in \eqref{eq:iou}
    and $M^{(t)}_{\text{alt}}$ in
    \eqref{eq:value_margin}. Metrics were collected using $1000$
    episodes and are averaged within timesteps.}
  \label{fig:metrics}
\end{figure}

To gain more information about how well the policy works in this environment, we
collected some metrics described in Section~\ref{sec:metrics}, as shown in
Fig.~\ref{fig:metrics}. As expected, at $t=0$, there is only one possible
state (the empty board), hence the two policies behave the same. At $t>0$,
however, the uncertainty in the state will cause the sets of actions taken by
the two policies to diverge. Also, the best action (according to the expected Q
values) is better by a significant margin than the others, particularly those of
the alternative policy.

Since the average IoU remains below $1$ in our experiments (as observable in
Fig.~\ref{fig:metrics}), for a significant number of possible observation-action
histories, an optimal action in a state of maximum probabality is not included
in the set of actions of the Q mixture-based policy.

$M_{\text{alt}}$ is close to $0.2$ throughout the episode in
Fig.~\ref{fig:metrics}, which is comparable as the advantage in average return
in Table~\ref{tab:average_returns} for the corresponding sense window size
$2\times2$.

%% file: conclusion.tex
\section{Conclusion}

We proposed a way to use policies solving the fully observable version of a
\pomdp{} to solve the \pomdp{} itself. Our method is applicable for all problems
where there is a solution for the fully observable version, and is especially
useful when the belief is multimodal making decisions based on only one state is
not enough.

We introduced a mathematical framework for our method and any further discussion
on it, and designed an appropriate benchmark to measure performance. We showed
that our policy consistently outperforms the benchmark policy that only uses the
state with maximum probability by $0.2$--$0.4$ in average return.

For this work to be applicable in practice, the effects of imperfect belief
estimation and also of the specific observation function on the performance of
the agent have to be examined to determine the class of problems this technique
is most applicable to. For example, when using a sampling-based belief
reconstruction method, the number of samples needed for optimal decision should
be analyzed.

We experimented in a small state space to be able to calculate the belief
analytically. To scale our approach to more real and usable applications, a
method of belief reconstruction will be needed.


%% file: paper.bbl
\begin{thebibliography}{10}
\providecommand{\url}[1]{#1}
\csname url@samestyle\endcsname
\providecommand{\newblock}{\relax}
\providecommand{\bibinfo}[2]{#2}
\providecommand{\BIBentrySTDinterwordspacing}{\spaceskip=0pt\relax}
\providecommand{\BIBentryALTinterwordstretchfactor}{4}
\providecommand{\BIBentryALTinterwordspacing}{\spaceskip=\fontdimen2\font plus
\BIBentryALTinterwordstretchfactor\fontdimen3\font minus
  \fontdimen4\font\relax}
\providecommand{\BIBforeignlanguage}[2]{{%
\expandafter\ifx\csname l@#1\endcsname\relax
\typeout{** WARNING: IEEEtran.bst: No hyphenation pattern has been}%
\typeout{** loaded for the language `#1'. Using the pattern for}%
\typeout{** the default language instead.}%
\else
\language=\csname l@#1\endcsname
\fi
#2}}
\providecommand{\BIBdecl}{\relax}
\BIBdecl

\bibitem{Sutton2018}
\BIBentryALTinterwordspacing
R.~S. Sutton and A.~G. Barto, \emph{Reinforcement learning: An introduction},
  2nd~ed.\hskip 1em plus 0.5em minus 0.4em\relax MIT press, 2018. [Online].
  Available: \url{http://incompleteideas.net/book/the-book-2nd.html}
\BIBentrySTDinterwordspacing

\bibitem{mnih2015}
\BIBentryALTinterwordspacing
V.~Mnih, K.~Kavukcuoglu, D.~Silver, A.~A. Rusu, J.~Veness, M.~G. Bellemare,
  A.~Graves, M.~Riedmiller, A.~K. Fidjeland, G.~Ostrovski \emph{et~al.},
  ``Human-level control through deep reinforcement learning,'' \emph{Nature},
  vol. 518, no. 7540, pp. 529--533, 2015. [Online]. Available:
  \url{https://doi.org/10.1038/nature14236}
\BIBentrySTDinterwordspacing

\bibitem{ppo}
\BIBentryALTinterwordspacing
J.~Schulman, F.~Wolski, P.~Dhariwal, A.~Radford, and O.~Klimov, ``Proximal
  policy optimization algorithms,'' \emph{CoRR}, vol. abs/1707.06347, 2017.
  [Online]. Available: \url{http://arxiv.org/abs/1707.06347}
\BIBentrySTDinterwordspacing

\bibitem{mpo}
\BIBentryALTinterwordspacing
A.~Abdolmaleki, J.~T. Springenberg, Y.~Tassa, R.~Munos, N.~Heess, and M.~A.
  Riedmiller, ``Maximum a posteriori policy optimisation,'' in \emph{6th
  International Conference on Learning Representations, {ICLR} 2018, Vancouver,
  BC, Canada, April 30 - May 3, 2018, Conference Track Proceedings}.\hskip 1em
  plus 0.5em minus 0.4em\relax OpenReview.net, 2018. [Online]. Available:
  \url{https://openreview.net/forum?id=S1ANxQW0b}
\BIBentrySTDinterwordspacing

\bibitem{pomdp_control}
\BIBentryALTinterwordspacing
E.~J. Sondik, ``The optimal control of partially observable markov processes
  over the infinite horizon: Discounted costs,'' \emph{Operations Research},
  vol.~26, no.~2, pp. 282--304, 1978. [Online]. Available:
  \url{http://www.jstor.org/stable/169635}
\BIBentrySTDinterwordspacing

\bibitem{Komorowski2018}
M.~Komorowski, L.~A. Celi, O.~Badawi, A.~C. Gordon, and A.~A. Faisal, ``The
  artificial intelligence clinician learns optimal treatment strategies for
  sepsis in intensive care,'' \emph{Nature Medicine}, vol.~24, no.~11, p. 1716,
  2018.

\bibitem{Mandel2014}
\BIBentryALTinterwordspacing
T.~Mandel, Y.~Liu, S.~Levine, E.~Brunskill, and Z.~Popovic, ``Offline policy
  evaluation across representations with applications to educational games,''
  in \emph{International conference on Autonomous Agents and Multi-Agent
  Systems, {AAMAS} '14, Paris, France, May 5-9, 2014}.\hskip 1em plus 0.5em
  minus 0.4em\relax International Foundation for Autonomous Agents and
  Multiagent Systems, 2014, pp. 1077--1084. [Online]. Available:
  \url{http://dl.acm.org/citation.cfm?id=2617417}
\BIBentrySTDinterwordspacing

\bibitem{Hauskrecht2000}
\BIBentryALTinterwordspacing
M.~Hauskrecht, ``Value-function approximations for partially observable markov
  decision processes,'' \emph{J. Artif. Intell. Res.}, vol.~13, pp. 33--94,
  2000. [Online]. Available: \url{https://doi.org/10.1613/jair.678}
\BIBentrySTDinterwordspacing

\bibitem{Markowitz2018}
\BIBentryALTinterwordspacing
J.~Markowitz, R.~W. Gardner, and A.~J. Llorens, ``On the complexity of
  reconnaissance blind chess,'' \emph{CoRR}, vol. abs/1811.03119, 2018.
  [Online]. Available: \url{http://arxiv.org/abs/1811.03119}
\BIBentrySTDinterwordspacing

\bibitem{newman2016reconnaissance}
A.~J. Newman, C.~L. Richardson, S.~M. Kain, P.~G. Stankiewicz, P.~R. Guseman,
  B.~A. Schreurs, and J.~A. Dunne, ``Reconnaissance blind multi-chess: an
  experimentation platform for isr sensor fusion and resource management,'' in
  \emph{Signal Processing, Sensor/Information Fusion, and Target Recognition
  XXV}, vol. 9842.\hskip 1em plus 0.5em minus 0.4em\relax International Society
  for Optics and Photonics, 2016, p. 984209.

\bibitem{rbc_conclusion}
\BIBentryALTinterwordspacing
R.~W. Gardner, C.~Lowman, C.~Richardson, A.~J. Llorens, J.~Markowitz,
  N.~Drenkow, A.~Newman, G.~Clark, G.~Perrotta, R.~Perrotta, T.~Highley,
  V.~Shcherbina, W.~Bernadoni, M.~Jordan, and A.~Asenov, ``The first
  international competition in machine reconnaissance blind chess,'' in
  \emph{NeurIPS 2019 Competition and Demonstration Track, 8-14 December 2019,
  Vancouver, Canada. Revised selected papers}, ser. Proceedings of Machine
  Learning Research, H.~J. Escalante and R.~Hadsell, Eds., vol. 123.\hskip 1em
  plus 0.5em minus 0.4em\relax {PMLR}, 2019, pp. 121--130. [Online]. Available:
  \url{http://proceedings.mlr.press/v123/gardner20a.html}
\BIBentrySTDinterwordspacing

\bibitem{openai_rubik}
\BIBentryALTinterwordspacing
OpenAI, I.~Akkaya, M.~Andrychowicz, M.~Chociej, M.~Litwin, B.~McGrew,
  A.~Petron, A.~Paino, M.~Plappert, G.~Powell, R.~Ribas, J.~Schneider,
  N.~Tezak, J.~Tworek, P.~Welinder, L.~Weng, Q.~Yuan, W.~Zaremba, and L.~Zhang,
  ``Solving rubik's cube with a robot hand,'' \emph{CoRR}, vol. abs/1910.07113,
  2019. [Online]. Available: \url{http://arxiv.org/abs/1910.07113}
\BIBentrySTDinterwordspacing

\bibitem{Kurniawati2008}
\BIBentryALTinterwordspacing
H.~Kurniawati, D.~Hsu, and W.~S. Lee, ``{SARSOP:} efficient point-based {POMDP}
  planning by approximating optimally reachable belief spaces,'' in
  \emph{Robotics: Science and Systems IV, Eidgen{\"{o}}ssische Technische
  Hochschule Z{\"{u}}rich, Zurich, Switzerland, June 25-28, 2008}, O.~Brock,
  J.~Trinkle, and F.~Ramos, Eds.\hskip 1em plus 0.5em minus 0.4em\relax The
  {MIT} Press, 2008. [Online]. Available:
  \url{http://www.roboticsproceedings.org/rss04/p9.html}
\BIBentrySTDinterwordspacing

\bibitem{Jaakkola1995}
\BIBentryALTinterwordspacing
T.~S. Jaakkola, S.~Singh, and M.~I. Jordan, ``Reinforcement learning algorithm
  for partially observable markov decision problems,'' in \emph{Advances in
  Neural Information Processing Systems 7, {[NIPS} Conference, Denver,
  Colorado, USA, 1994]}, G.~Tesauro, D.~S. Touretzky, and T.~K. Leen,
  Eds.\hskip 1em plus 0.5em minus 0.4em\relax {MIT} Press, 1994, pp. 345--352.
  [Online]. Available:
  \href{http://papers.nips.cc/paper/951-reinforcement-learning-algorithm-for-partially-observable-markov-decision-problems}{http://papers.nips.cc/paper/951-reinforcement-learning-algorithm-for-partially-observable-markov-decision-problems}
\BIBentrySTDinterwordspacing

\bibitem{wierstra2007}
D.~Wierstra, A.~Förster, J.~Peters, and J.~Schmidhuber, ``Solving deep memory
  pomdps with recurrent policy gradients,'' vol. 4668, 09 2007, pp. 697--706.

\bibitem{Hausknecht2015}
M.~Hausknecht and P.~Stone, ``Deep recurrent q-learning for partially
  observable mdps,'' in \emph{2015 AAAI Fall Symposium Series}, Jul. 2015.

\bibitem{Clark2021}
\BIBentryALTinterwordspacing
G.~Clark, ``Deep synoptic monte-carlo planning in reconnaissance blind chess,''
  in \emph{Advances in Neural Information Processing Systems 34: Annual
  Conference on Neural Information Processing Systems 2021, NeurIPS 2021,
  December 6-14, 2021, virtual}, M.~Ranzato, A.~Beygelzimer, Y.~N. Dauphin,
  P.~Liang, and J.~W. Vaughan, Eds., 2021, pp. 4106--4119. [Online]. Available:
  \url{https://proceedings.neurips.cc/paper/2021/hash/215a71a12769b056c3c32e7299f1c5ed-Abstract.html}
\BIBentrySTDinterwordspacing

\bibitem{Blowitski2021}
K.~Blowitski and T.~Highley, ``Checkpoint variations for deep q-learning in
  reconnaissance blind chess,'' \emph{J. Comput. Sci. Coll.}, vol.~37, no.~3,
  p. 81–88, oct 2021.

\bibitem{Highley2020}
T.~Highley, B.~Funk, and L.~Okin, ``Dealing with uncertainty: A piecewisegrid
  agent for reconnaissance blind chess,'' \emph{J. Comput. Sci. Coll.},
  vol.~35, no.~8, p. 156–165, apr 2020.

\bibitem{Guo2020}
\BIBentryALTinterwordspacing
Z.~Guo, X.~Wang, S.~Qi, T.~Qian, and J.~Zhang, ``Heuristic sensing: An
  uncertainty exploration method in imperfect information games,''
  \emph{Complexity}, vol. 2020, pp. 8\,815\,770:1--8\,815\,770:9, 2020.
  [Online]. Available: \url{https://doi.org/10.1155/2020/8815770}
\BIBentrySTDinterwordspacing

\end{thebibliography}
